\documentclass{sig-alternate}

\usepackage{graphicx}

\begin{document}

%
\conferenceinfo{KDD CUP Workshop}{2011 USA}

\title{Efficient Multicore Collaborative Filtering}
%
%
%
%
%

\numberofauthors{6} 
%
\author{
%
%
\alignauthor
Yao Wu \\
       \affaddr{Institute of Automation}\\
       \affaddr{Chinese Academy of Sciences}\\
       \affaddr{Beijing 100190, China}\\
       \email{wuyao@nlpr.ia.ac.cn}\\
\alignauthor
Qiang Yan \\
       \affaddr{Institute of Automation}\\
       \affaddr{Chinese Academy of Sciences}\\
       \affaddr{Beijing 100190, China}\\
       \email{scmyyan@gmail.com}\\
\alignauthor
Danny Bickson\\
       \affaddr{Carnegie Mellon University}\\
       \affaddr{5000 Forbes Ave}\\
       \affaddr{Pittsburgh, PA}\\
       \email{bickson@cs.cmu.edu}\\
\and \\
\alignauthor
Yucheng Low\\
       \affaddr{Carnegie Mellon\ University}\\
       \affaddr{5000 Forbes Ave}\\
       \affaddr{Pittsburgh, PA}\\
       \email{ylow@cs.cmu.edu}
\alignauthor
Qing Yang \\
      \affaddr{Institute of Automation}\\
      \affaddr{Chinese Academy of Sciences}\\
       \affaddr{Beijing 100190, China}\\
       \email{qyang@nlpr.ia.ac.cn}   \\
}

\maketitle
\begin{abstract}
This paper describes the solution method taken by LeBuSiShu team for track1 in ACM KDD CUP 2011 contest (resulting in the 5th place). We identified two main challenges: the unique item taxonomy characteristics as well as the large data set size.

To handle the item taxonomy, we present a novel method  called Matrix Factorization Item Taxonomy Regularization (MFITR). MFITR obtained the 2nd best prediction result out of more then ten implemented algorithms. 

For rapidly computing multiple solutions of various algorithms, we have implemented an open source parallel collaborative filtering library on top of  the GraphLab machine learning framework. We report some preliminary performance results obtained using the BlackLight supercomputer. \end{abstract}


\terms{Machine learning, data mining}


\section{Introduction}
The task in the ACM KDD\ CUP\ track1 was to predict music ratings using a real dataset obtained from the Yahoo! music service. A full description of the dataset is given in \cite{KDD-dataset}. There are two main factors which make the prediction task challenging. Firstly, the magnitude of the dataset is rather large: there are 1,000,990 users, 624,961 music items (songs) and 262,810,175 user ratings, spanning over 6649 time bins. For data of this magnitude, commonly used mathematical software like Matlab can not be efficiently deployed. Secondly, the data includes additional features such as the time when the user ratings were recorded as well as the hierarchy of rated items to genres (each rated song can belong to one or more genre),  album and artist.

In this paper we describe how we handled the two challenges described above. Section \ref{sec:algos} outlines the theoretical algorithms used for computing the prediction. Section \ref{sec:mfitr} explains how those algorithms where adapted to the KDD CUP contest, namely accounting for hierarchy of data items. Section \ref{sec:graphlab} discusses our efficient custom parallel implementation on top of the GraphLab machine learning framework, that was used to rapidly fine-tune multiple algorithm parameters, including report of performance results. We conclude in Section 5.

As an additional contribution, we release open source code of many of the implemented algorithms as part of GraphLab's collaborative filtering library - available from\\ 
{\tt http://graphlab.org/}.

\section{Algorithms}
\label{sec:algos}
Inspired by the Bellkor team's algorithm which won the Netflix contest \cite{Bellkor}, we deployed an ensemble method, combining a collection of collaborative filtering algorithms while blending the solutions together. The ensemble comprises of 12 methods listed in Table \ref{algorithms_table}, of which the last two are novel. In the rest of this section we describe the implemented algorithms in more detail.
\begin{table}[h!]
\begin{tabular}{|c|l|l|}\hline
1 &  Item-kNN \cite{SGD} & Neighborhood based\\ \hline
2 & ALS \cite{ALS} & Alternating least squares \\\hline
3 & wALS \cite{wALS} & Weighted alternating least squares \\\hline
4 & BPTF \cite{SDM10,ICML08} & Bayesian prob. tensor factorization\\\hline
5 & SGD \cite{SGD,SGD2} & Stochastic gradient descent \\\hline
6 & SVD++ \cite{SVD++} & SVD++ algorithm\\\hline
7 & time-kNN \cite{Koren2009} & Time aware neighborhood model \\\hline
8 & time-SGD \cite{Koren2009} & Time aware SGD \\\hline
9 & time-SVD++ \cite{Koren2009} & Time aware SVD++ \\\hline
10 & Random-forest \cite{Breiman_2001} & Random forest \\\hline\hline
11 & {\bf MFITR}  & MF  item taxonomy regularization\\\hline
12 &{\bf time-MFITR}  & MF item taxonomy regularization,\\
& & time aware\\\hline
\end{tabular}
\caption{Different algorithms implemented. The last two are our novel contribution.}
\label{algorithms_table}
\end{table}
\newpage
\subsection{Neighborhood models}
An item-based neighborhood approach predicts the rating $r_{ui}$ of a user $u$ for a new item $i$, using the rating of the user $u$ gave to the items which are similar to $i$. We choose the Adjusted 
Cosine (AC) similarity to measure the similarity $w_{ij}$ between item $i$ and $j$.
\small
\begin{displaymath}
w_{ij}=\frac{\sum_{u\in U_{ij}} (r_{ui}-\overline{r}_u)(r_{uj}-\overline{r}_u)}
       {\sqrt{\sum_{u\in U_{ij}} (r_{ui}-\overline{r}_u)^2
       \sum_{u\in U_{ij}} (r_{uj}-\overline{r}_u)^2}}\,.
\end{displaymath}
\normalsize

Here, $U_{ij}$ denotes the users who have rated both item $i$ and $j$.
Based on the similarity, for every item $i$, we can compute the neighborhood $N_i$ which contain the $K$ items most similar to $i$. Then we can predict $\widehat{r}_{ui}$ based on the items in both 
$N_i$ and $R_u$ which is the set of ratings made by user $u$:
\begin{displaymath}
\widehat{r}_{ui} = \frac{\sum_{j\in R_u \cap N_i}w_{ij}r_{uj}}
                 {\sum_{j\in R_u \cap N_i}\lvert w_{ij}\rvert}\,.
\end{displaymath}
Here, $\cap$ denotes the intersection of two sets.

To address the computational challenges arising from the huge number of items, we split the items into $N$ parts. For each $i$th iteration, we only need to compute the neighbors of the items in the $i$th part. In our experiments, we set $N=300$, thus matrix $M_{I\times J}$ fitted into a 8GB memory computer, here $J=I/N$ and $I$ is the number of items. This method can be easily parllelized.

\subsection{ALS and BPTF}
Alternating least squares \cite{ALS} is a simple matrix factorization algorithm. The non-zero rating of item form a matrix $A$ of size $M \times N$, where $M$ number of users and $N$ in the number of items.
The matrix $A$ is decomposed into two low rank matrices $A \approx U*V$ where $U$ is of size $M \times D$ and $V$ is $D \times N$. Starting from an initial guess, each iteration first fixes $U$ and computes
$V$ using a least squares procedure, then fixes $V$ and computes $U$ using the same least square procedure. The rating is computed as a vector product of the matching user and item feature vectors: 
\[ \widehat{r}_{ui}(t) = \sum_{j=1}^D U_{u,j} V_{j,i}\,. \]
ALS model can be extended to the tensor case where time information is included with the rating. 
Bayesian probabilistic tensor factorization (BPTF) \cite{SDM10} is a Markov Chain Monte Carlo method, where on top of the least squares step, sampling from the hyperpriors of $U,V$ is added.

\subsection{SGD}
Matrix Factorization methods have demonstrated superior performance vs. neighborhood based
models \cite{SVD++}.
Matrix factorization models map both users and items to a joint latent factor space of dimension $D$, such the user-item interactions are modeled as inner products in that space. Each item $i$ and user 
$u$ is associated with a $D$-dimensional latent feature vector $q_i$ and $p_u$ respectively. Thus predicted rating is computed by:
\begin{displaymath}
          \widehat{r}_{ui} = \mu + b_i + b_u + q_i^Tp_u\,.
\end{displaymath}

        The parameters $b_i$, $b_u$, $q_i$ and $p_u$ are learned by minimizing a certain loss function based on the $(u,i)$ pairs in the set of observed ratings $O$:
\begin{equation}
          \min \sum_{(u,i)\in O} (r_{ui}-\widehat{r}_{ui})^2+ \lambda(b_i^2+b_u^2+\lVert q_i\rVert^2+\lVert p_u\rVert^2)\label{SVDcost}
\end{equation}
        where $\lVert .\rVert^2$ denotes the Frobenius 2-norm and the positive constant $\lambda$ controls the extent of regularization and it is determined by cross validation. 
We used stochastic gradient descent optimization to minimize the loss function \eqref{SVDcost}. The complexity of each iteration is linear in the number of ratings. 
\subsection{SVD++}
Implicit feedback can improve the prediction accuracy since it provides an additional indication of user preferences. SVD++ \cite{SVD++} is an extension of the linear model of \eqref{SVDcost}. For each item $i$, we 
add an additional latent
factor  $y_i$. Thus, the latent factor vector of each user $u$ can be characterized by the set of items the user have rated. The exact model is as follows:
\begin{displaymath}
          \widehat{r}_{ui} = \mu + b_i + b_u + q_i^T(p_u+\lvert R_u\rvert^{-1/2}\sum_{j\in R_u}y_j)\,.
\end{displaymath}

  Similarly, we can learn the parameters $b_i$, $b_u$, $q_i$, $p_u$ and $y_i$ using stochastic gradient descent optimization to minimizing the quadratic loss function. Again, the complexity per iteration is linear in the number of ratings.

\label{sec:time}
\subsection{Time-aware neighborhood models}
In the time-aware cf model, each rating $r_{ui}$ is associated with a time stamp $t_{ui}$, which indicates the time when the rating was observed. However, a rating $r_{ui}$ observed 3 years ago is less important as a rating $r_{uj}$ taken 3 days ago, when used to predict the current ratings.

Following \cite{Koren2009}, we define a time-decay function to model this effect:
$$f_{ui}(t)=e^{-\beta(t-t_{ui})}\,,$$
where $\beta\ge 0$ controls the decaying rate. When $\beta=0$, we don't consider the temporal effects.

We can incorporate the temporal effect into the neighborhood models as follows:
\begin{displaymath}
\widehat{r}_{ui}(t) = \frac{\sum_{j\in R_u \cap N_i}f_{ui}(t)w_{ij}r_{uj}}
                 {\sum_{j\in R_u \cap N_i}f_{ui}(t)\lvert w_{ij}\rvert}\,.
\end{displaymath}

In our experiments, we found that setting $\beta=0.08$ gave the best performance. Overall, time-aware neighborhood model achieved a significantly better result than time-independent neighborhood 
model.

\subsection{Time-aware matrix factorization}
Like in \cite{Bellkor}, we model the temporal effect into the Matrix Factorization by allowing the parameters vary with different time. In particular, we quantize the user-based time effect by days $T_1,T_2,...,T_N$ and add a item-time-bin bias to the equation. The predicted rating is computed
as follows:\begin{displaymath}
          \widehat{r}_{ui}(t) = \mu + b_i + b_u + b_{u,t} + b_{i,Bin(t)}+
              q_i^T(p_u+\lvert R_u\rvert^{-1/2}\sum_{j\in R_u}y_j)\,.
\end{displaymath}
Here, $b=Bin(t), b=0,1,...,N$, we choose $N=30$ as the number of time bins. $b_{u,t}$ is a 2-dimensional array factorized by $b_{u,t}=x_{u}^Tz_{t}$ to cut memory cost:
\begin{displaymath}
          \widehat{r}_{ui}(t) = \mu + b_i + b_u + x_{u}^Tz_{t} + b_{i,Bin(t)}+
              q_i^T(p_u+\lvert R_u\rvert^{-1/2}\sum_{j\in R_u}y_j)\,.
\end{displaymath}

Hence, we could extend the time-independent loss function to the following form:
\begin{eqnarray*}
\min \sum_{(u,i,t)\in O} (r_{uit}- \widehat{r}_{ui}(t))^2
     +\lambda_1(b_i^2+b_u^2+b_{i,Bin(t)}^2)+\\+\lambda_2(\lVert q_i\rVert^2+\lVert p_u\rVert^2+\sum_{j \in R_u}\lVert y_j\rVert^2)
     +\lambda_3(\lVert x_u\rVert^2+\lVert z_t\rVert^2) \,.
\end{eqnarray*}
Similarly, the model described in the section \ref{sec:mfitr} can be extended to a time-aware version.
\begin{displaymath}
          \widehat{r}_{ui}(t) = \mu + b_i + b_u + b_{ar} + x_{u}^Tz_{t} + b_{i,Bin(t)} + (q_i + q_{ar})^Tp_u\,.
\end{displaymath}

\subsection{Random Forests}
The above techniques largely make use of only rating information. To utilize the remaining information such as ``album'' and ``artist info'', we used random forests to perform regression of all item features.
We consider this prediction task as to estimate a function $\bold x_{ui}\to r_{ui}$. Here, $\bold x_{ui}$ is a D-dimensional vector which maps to the features of a $(u,i)$ pair and $r_{ui}$ denotes the rating. In our experiments, the features included the user id, item id, artist, album and genres. We use random forests \cite{Breiman_2001} to regress the features. The single model did not perform as well as traditional Collaborative Filtering algorithms (obtained RMSE 26 on validation dataset), but we found it can improve our final solution after blending with other models (obtaining an improvement of 0.08 on the leaderboard).

\subsection{Blending multiple solutions}
\label{sec:blending}
A lesson learned from the Netflix contest is that the combination of different algorithms can lead to significant performance improvement over individual algorithms \cite{Bigchaos09}. We blended our multiple predictors based on a linear regression model. We use the validation set to compute the optimal $M$-dimensional linear combination weights $\bold w$, where $M$ is the number of blending models. 

First, we trained each model $m_i$ independently based on the training set and get the predictions $\bold x_i$ of the $N$ ratings in validation set, where $x_i$ is
a $N$-dimensional vector. The target value for the $N$ data points are $\bold y$, the weights $\bold w$ can be obtained by solving a least squares problem. In our solution, we used ridge regression:
$$\bold w = (\bold X^T\bold X + \lambda \bold I)^{-1}\bold X\bold y\,,$$
where $\bold I$ denotes the identity matrix and the regularization $\lambda$ is determined by cross validation \cite {JahreKDD10}.

Next, every model is trained again using the same initialization parameters, but training is now performed using both the training set and the validation set. Finally, test set predictions of each model $m_i$ is computed ($\hat{\bold x}_i$) and the final prediction is obtained:
$$\widehat{\bold r} = \widehat{\bold X}^T\bold w\,.$$

\section{MFITR:\ Matrix factorization with item taxonomy regularization}
\label{sec:mfitr}
A unique property of the KDD data, is that tracks, albums, artists and genres form a hierarchy; where each track belongs to an album, each albums belongs to an artist, 
and both are tagged by genres \cite{KDD-dataset}. We propose MFITR, a novel method to utilize  item taxonomy information to improve  prediction accuracy.

To capture the hierarchy of items, we construct a graph between tracks, albums and artists. (We did not use genre information.) The method is different from the traditional matrix 
factorization since we model the item hierarchy as a regularization term to constrain the matrix factorization computation, as explained next. A closely related approach is recently proposed in the  social network domain \cite{Ma2011}. 

We assume an object $i$ is a parent of another object $j$ if there exists a hierarchic relationship between them and $j$ belongs to $i$, meanwhile, $j$ can be seen as a child of $i$. The root 
nodes of the hierarchic relationship are artists, and the children nodes are albums. Tracks belonging to an album are children of the album.  This item hierarchy is therefore a graph and we denote the parent set of item $i$ as $P_i$ and the child set as $C_i$.

If a user $u$ have given an artist $a_i$ a rating 100 and given another artist $a_j$ a rating 0, we can intuitively know that $u$ will like tracks and albums of $a_i$ more than that of $a_j$. 
Based on this intuition, we propose a model based on matrix factorization. The prediction is computed by: 

$$\widehat{r}_{ui} = \mu + b_i + b_u + b_{a} + (q_i + q_{a})^Tp_u\,.$$

We use $b_{a}$ as the bias of the artist $a$ and $q_{a}^Tp_u$ as the user feature vector for artist $a$, the performer of music item $i$.
Furthrmore, to support different ratings of tracks in the same album, we propose a more advanced model:
\small
\begin{eqnarray*}
\min \sum_{(u,i)\in O} (r_{ui}-\mu - b_i - b_u - b_{a} - (q_i + q_{a})^Tp_u)^2+ \\
     +\lambda_1(b_i^2+b_u^2+b_{a}^2)+\lambda_2(\lVert q_i\rVert^2+\lVert p_u\rVert^2+||q_{a}||^{2})+\\
     +\lambda_3\sum_{i}\sum_{j\in P_i}w_{ij}\lVert q_i-q_j\rVert^2
     +\lambda_4\sum_{i}\sum_{j\in C_i}w_{ij}\lVert q_i-q_j\rVert^2\,.
\end{eqnarray*} 
\normalsize
Here, $w_{ij}$ is the similarity between $i$ and $j$, computed as in the neighborhood model. If $i$ and $j$ are similar, the distance between $q_i$ and $q_j$ shouldn't be large.
Table \ref{imftr_notations} summarizes the notations used in IMFTR.

\begin{table}[h!]
\centering{
\begin{tabular}{|c|l|}\hline
$r_{ui}$ & Observed rating for item $i$ by user $u$\\ \hline
$\widehat{r}_{ui}$ & Predicted rating for item $i$ by user $u$\\ \hline
$b_i$ & Bias for item $i$ \\\hline
$b_u$ & Bias for user $u$ \\\hline
$b_a$ & Bias for artist $a$ \\\hline
$\mu$ & Mean rating \\\hline
$q_i$ & Feature vector for item $i$ \\\hline
$q_u$ & Feature vector for user $u$ \\\hline
$q_a$ & Feature vector for artist $a$ \\\hline
$\lambda_1$ & Weighting factor for biases \\\hline
$\lambda_2$ & Weighting factor for feature vectors \\\hline
$\lambda_3$ & Weighting factor for child similarity \\ \hline
$\lambda_4$ & Weighting factor for parent similarity \\ \hline
$w_{ij}$ & neighborhood similarity between objects $i$ and $j$\\\hline
$C_i$ & Graph children set of object $i$ \\ \hline
$P_i$ & Graph parent set of object $i$ \\ \hline
\end{tabular}
}
\caption{MFITR Notations}
\label{imftr_notations}
\end{table}

The MFITR cost function is composed of four terms. The first term minimizes the Euclidean distance between the observed and predicted rating,
where biases of user item and artist are taken into account. The second term is a regularization term of the biases to prevent over-fitting. 
The third and fourth term enforce similarity between tracks in the same album and between albums of the same artist, when they are rated
closely by the neighborhood model. 

An advantage of this approach is that the similarity between parent and children can
propagate indirectly in the learning phase \cite{Ma2011}. For example, two tracks $t_a$ and $t_b$  from the same album $l$  have no direct relationship between them. However, since they have the same 
parent $l$,
the distance between $q_{t_a}$ and $q_{t_b}$ is actually minimized indirectly when the distances $w_{l,t_a}\lVert q_{l}-q_{t_a}\rVert^2$ and $w_{l,t_b}\lVert q_{l}-q_{t_b}\rVert^2$ are minimized.

An immediate extension we implemented is to add time information into the cost function as done in time-SVD++. We call this variant time-MFITR.
As shown in the next section, time-MFITR has very good performance on KDD data. 

\section{Efficient multicore implementation}
\label{sec:graphlab}
A majority of the algorithms described where implemented on top of the  GraphLab parallel machine learning framework \cite{uaigraphlab}. We selected GraphLab since it allowed us for rapid prototyping and testing of multiple CF algorithms. The following algorithms where implemented: ALS, weighted-ALS (wALS), SVD++, PMF, BPTF and SGD. Since we used multiple algorithms, where each algorithm had multiple tunable parameters that needed to be adjusted, an efficient parallel solution was essential for rapidly improving our model. All of the above algorithms are open sourced as part of the GraphLab collaborative filtering library:\\ {\tt http://graphlab.org/}.

We have utilized our own cluster (several AMD Opteron 8387 4-8 core machines, 2.7Ghz, 16-64GB memory) as well as the BlackLight \cite{BlackLight} supercomputer (SGI UV 1000 NUMA shared-memory system comprising 256 blades. Each blade holds 2 Intel Xeon X7560 Nehalem 2.27 Ghz eight-core processors, for a total of 4096 cores.) Overall, we estimate that we have used around 10,000 cpu hours on our clusters and 10,000 cpu hours on BlackLight.
Each algorithm was run in parallel using 8-32 cores using line search for each tunable parameter. Each of those runs was repeated twice: with and without validation data used for training, as explained in Section \ref{sec:blending}.

\subsection{Performance results}
\begin{figure}
\centering{
\includegraphics[scale=0.35]{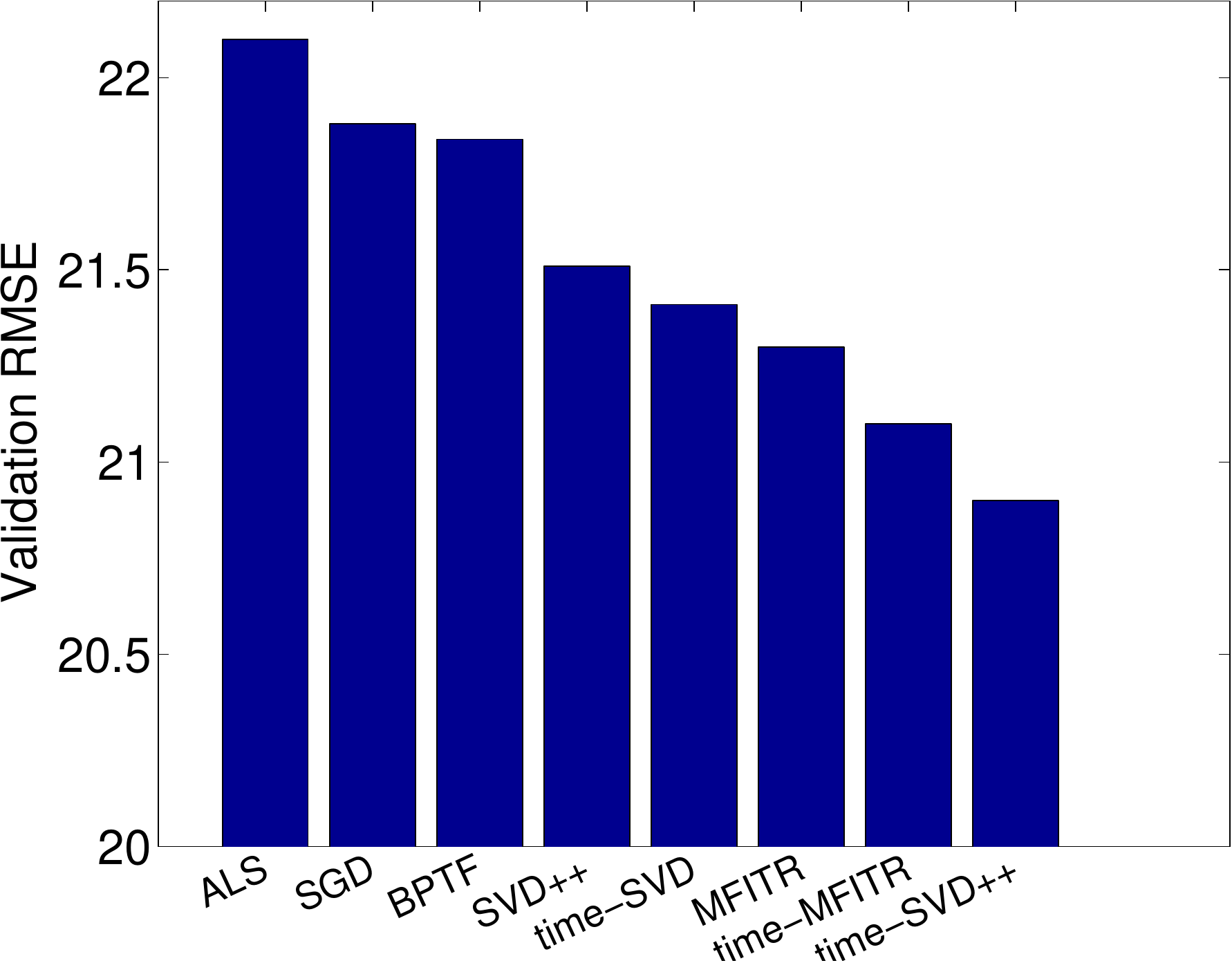}
\caption{RMSE of the different algorithms on the validation data.}
\label{rmseplot}
}
\end{figure}

Table \ref{maintable} lists the different tunable parameters we tested for each algorithm, and the optimized setting we found.
As a baseline for performance, we measure RMSE (root mean square error) on the validation dataset. 
The most effective single algorithm is time-SVD++ which obtained RMSE\ of 20.90 on the validation data. The second most effective single algorithm is our novel time-MFITR algorithm which obtained RMSE\ of 21.10 on the validation data.
Note that while wALS obtained the best performance, it did overfit and gave worse performance on the actual test data. A summary of the results is given in Figure \ref{rmseplot}.
\

\begin{figure}[h!]
\centering{
\includegraphics[scale=0.35]{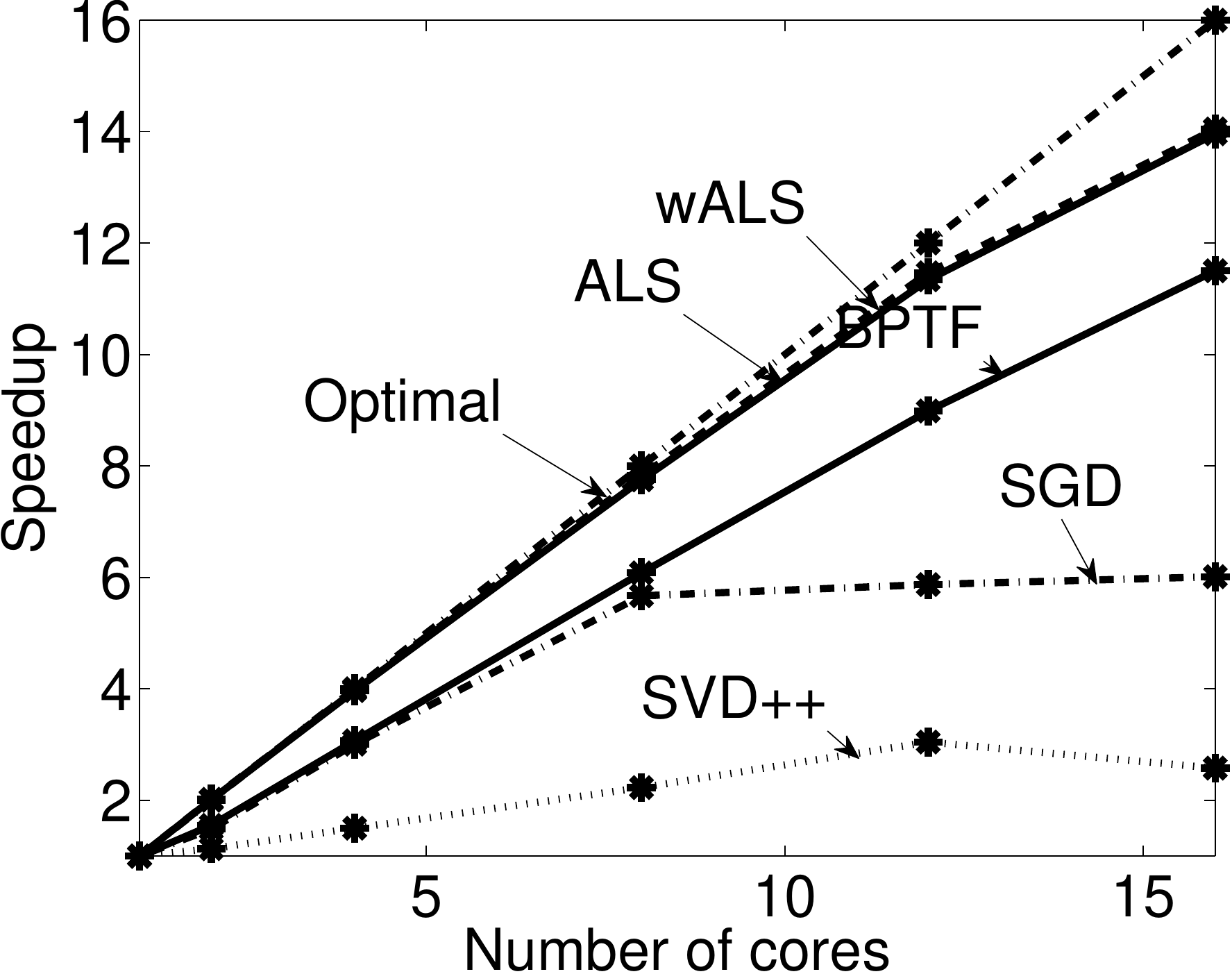}
\caption{Speedup of Graphlab using KDD data on BlackLight with up to 16 cores.}
\label{speedup}
}
\end{figure}

Regarding performance of the parallel implementation. Figure \ref{speedup} shows the speedup of several algorithms using BlackLight. Similar results were obtained also on the Linux cluster and will not be repeated here. 
Speedup is defined using the baseline of a single CPU run. For wALS,ALS we obtain an almost optimal speedup of x14 on 16 cores. BPTF performs slightly less since it has a sampling step after each iteration which is serial
and slows the algorithm a little. SGD, SVD++ performance is worse - with a speedup of about x6 and x3, respectively. The reason is that we deploy a locking mechanism to prevent users to update the same item  feature vector concurrently.
  
\begin{figure}[h!]
\centering{
\includegraphics[scale=0.35,clip=true]{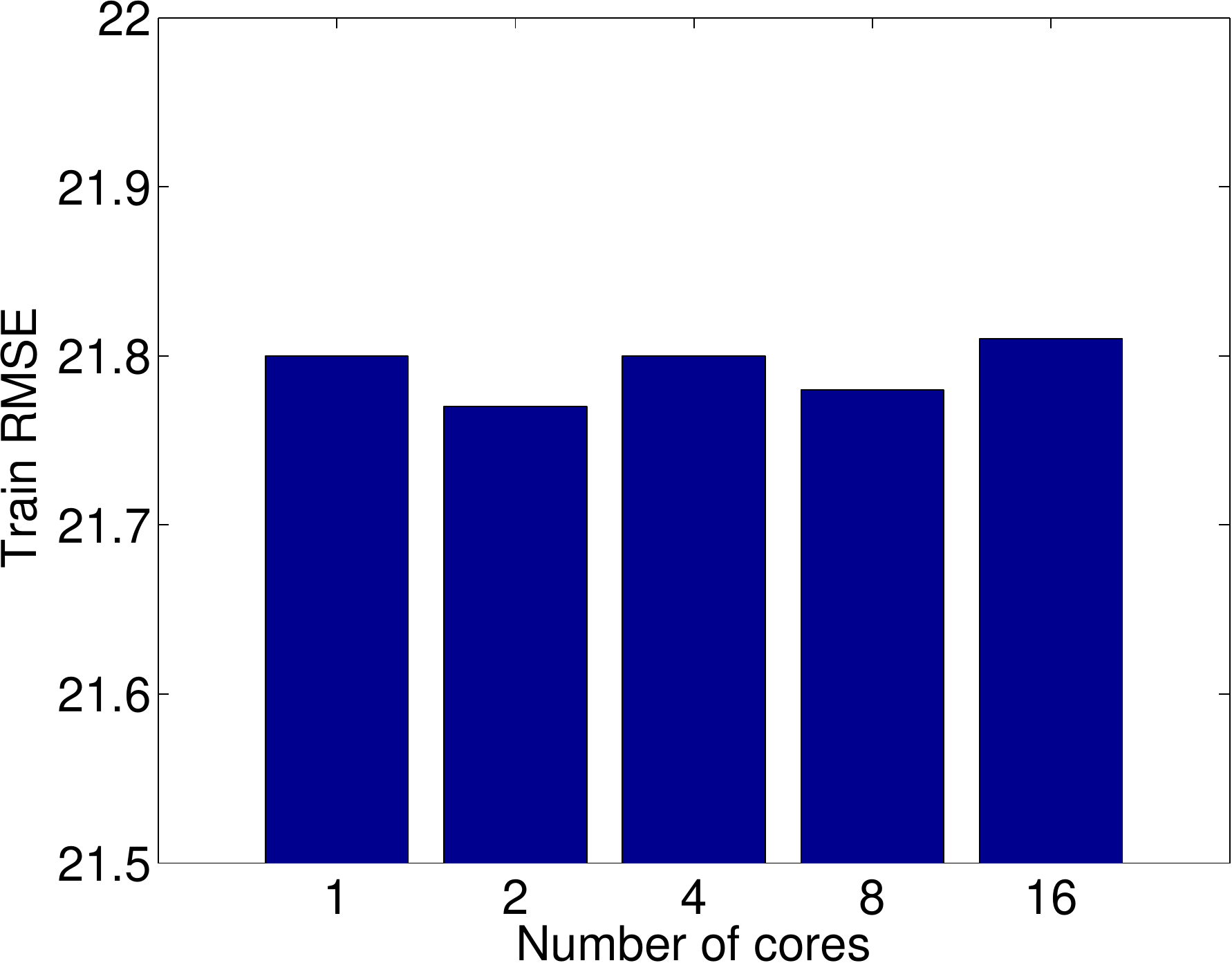}
\caption{Accuracy of SGD validation RMSE using different number of cores. Variations are up to 0.1\%.}
\label{accuracy}
}
\end{figure}

Regarding accuracy of the parallel computation vs. an equivalent serial result. Figure \ref{accuracy} examines the validation RMSE of 5 iterations the SGD algorithm (D=50) using different number of cores 1-16 on BlackLight.
Because of the parallel implementation there are slight variations in accuracy.
However, variations are not more than 0.1\% of the serial result. Similar behavior was observed for the other algorithms.

\begin{figure}[h!]
\centering{
\includegraphics[scale=0.35,clip=true]{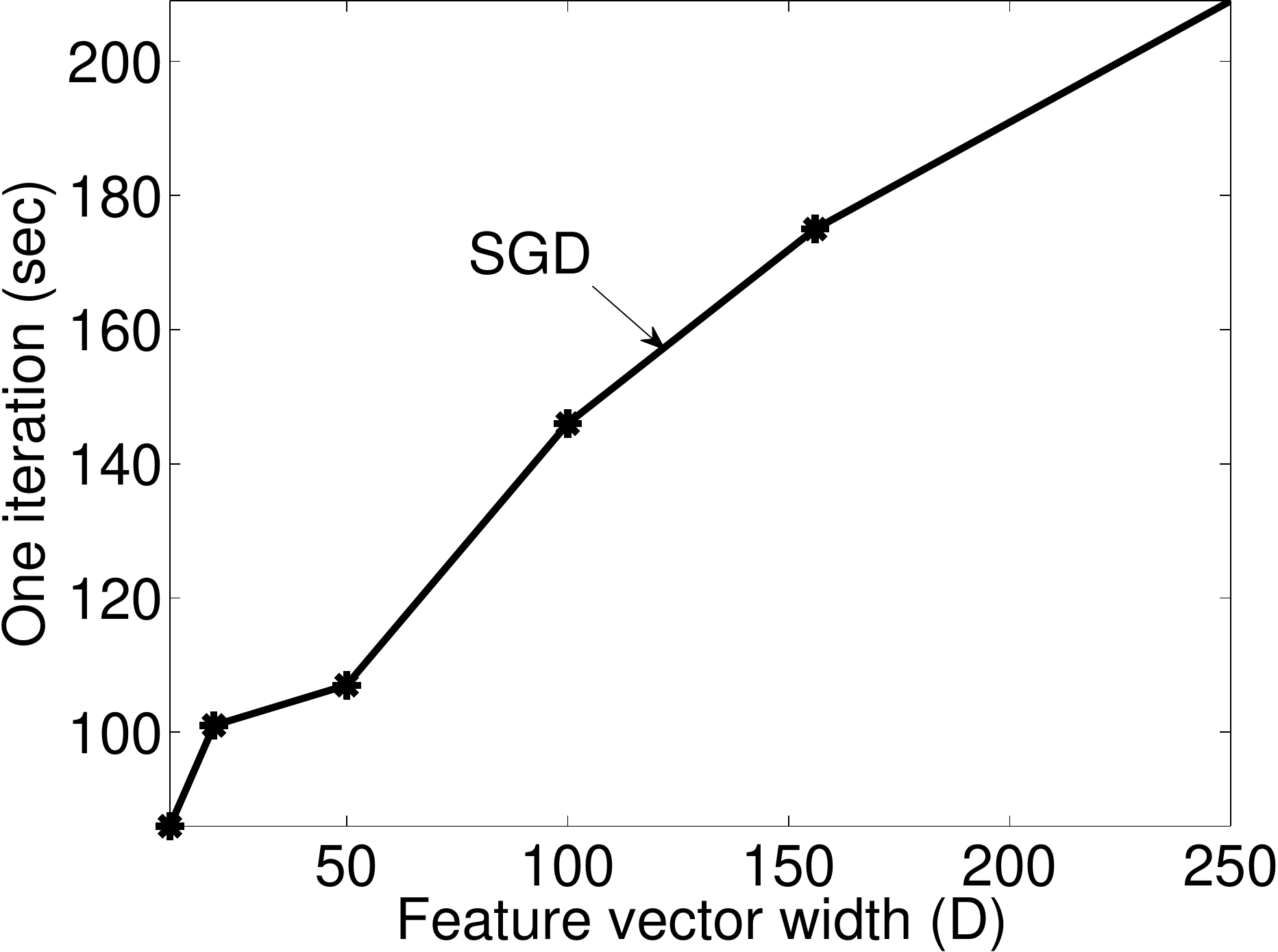}
\caption{SGD iteration time vs. feature vector width (D).}
\label{sgdplot}
}
\end{figure}

Another interesting question we looked at is how does computation scale with the length of the feature vector. Figure \ref{sgdplot} shows the good scaling of SGD algorithm. This scaling was also observed for SVD++. Both algorithms performance is almost linear with the number of features. ALS, wALS and BPTF all perform
matrix inversion as part of the update rule and thus the scaling is less good. 

\begin{figure}[h!]
\centering{
\includegraphics[scale=0.35,clip=true]{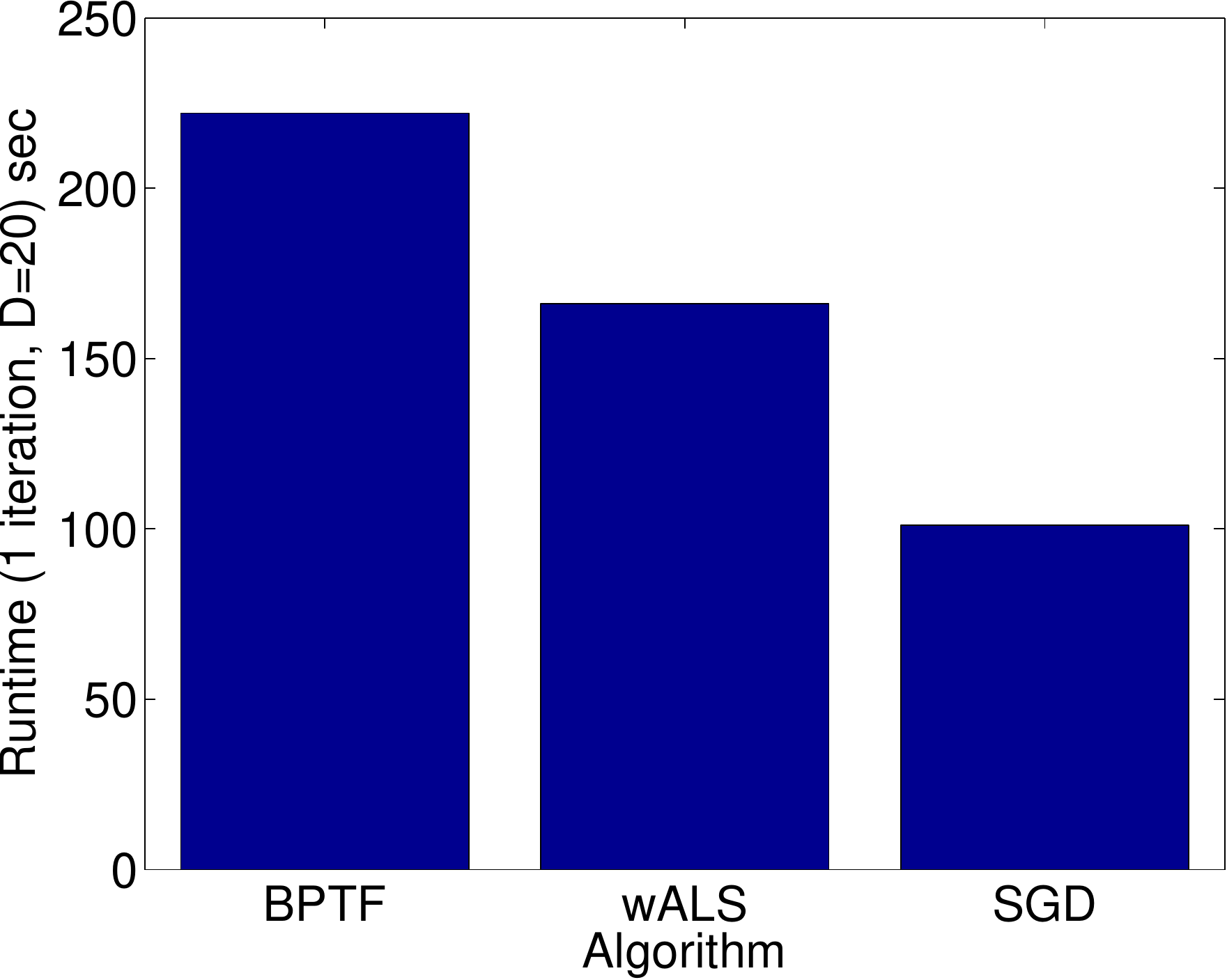}
\caption{Running time of a single iteration with 16 cores.}
\label{runtimeplot}
}
\end{figure}

Figure \ref{runtimeplot} depicts running time of a single iteration of several algorithms, on 16 cores with $D=20$. SGD and SVD++ (not shown, but has similar running time as SGD) are the fastest algorithms per iteration. But from the other hand, it is more difficult to make them work efficiently in parallel. 

\begin{table*}[h!]
\centering{
\begin{tabular}{|c|l|l|l|}\hline
NO & Algorithm & Tunable parameters &\ RMSE on validation\\\hline \hline 1& Neighborhood model& Adjusted cosine (AC)\ similarity & 23.34\\ \hline
2 & ALS & $\lambda$=regularization (1)&\\
& & $D$=feature vector width (120)&\\
& & $r$=number of iterations (50)&22.01\\\hline
3 & wALS & $\lambda$=regularization (1)&\\
& & $D$=feature vector width (120)&\\
& & $r$=number of iterations50& 18.87\\\hline
4 &BPTF & $b$=burn-in period (5)&\\
& & $D$=feature width (20) & \\
& & $s$=scaling of paramters (100)&\\
& & $r$=number of iterations (100)&21.84\\\hline
5 & SGD&  $q$ =learning rate decay (0.95)&\\
 & &$\gamma$=learning rate (5e-4) & \\
  & &$\lambda$=weight factor (1e-4) & \\
    & &$r$=iterations (50) & 21.92\\
   & &$r$=iterations (100) & 21.88\\ \hline
6 & SVD++ & $\gamma$=learning rate (5e-4)&\\
& &  $q$ =learning rate decay (0.95)&\\
& &$\lambda$=weight factor (1e-4) & \\
& &$\lambda$=weight factor (1e-4) & \\
& &$D$=features (50) & 21.65\\
& &$D$=features (100) & 21.59\\ \hline
7& Time aware neighborhood& $\beta$=time decay ($0.08$)& 22.7 \\\hline
8&time-SVD& $\lambda_1$ = 1e-4& \\
& & $\lambda_2,\lambda_3$=5e-4& \\
& & $\lambda_2$=5e-4& \\
 & &$\gamma$=learning rate (1e-4) & \\
 & &  $q$ =learning rate decay (0.95)&\\
 & &$D$=features (50) & 21.45\\
& &$D$=features (100) & 21.41\\ \hline
9&time-SVD++& $\lambda_1$ = 1e-5& \\
& & $\lambda_2$=1e-4& \\
& & $\lambda_3$=3e-4& \\
 & &$\gamma$=learning rate (5e-5) & \\
 & &  $q$ =learning rate decay (0.95)&\\
 & &$D$=features (50) & 20.97\\
& &$D$=features (100) & 20.90\\ \hline
10&MFITR& $\lambda_1$ = 1e-5& \\
& & $\lambda_2$=1e-4& \\
& & $\lambda_3$=1e-3& \\
& & $\lambda_4$=1e-3& \\
 & &$\gamma$=learning rate (8e-5) & \\
 & &  $q$ =learning rate decay (0.95)&\\
 & &$D$=features (50) & 21.39\\
& &$D$=features (100) & 21.30\\ \hline
11&time-MFITR& $\lambda_1$ = 1e-5& \\
& & $\lambda_2$=1e-4& \\
& & $\lambda_3$=1e-3& \\
& & $\lambda_4$=1e-3& \\
& & $\lambda_5$=1e-3& \\
 & &$\gamma$=learning rate (8e-5) & \\
 & &  $q$ =learning rate decay (0.95)&\\
 & &$D$=features (50) & 21.10\\ \hline
12&Random forest &  & 26.0\\
\hline\hline

- & Session based post-processing &Neighborhood model&23.34=>23.02\\
& & SGD &21.92=>21.77\\
& & SVD++ &21.65=>21.51\\\hline
- & Blending all together & & 19.90 \\ \hline
\end{tabular}
\caption{Main results on the validation data obtained using the different algorithms.}
\label{maintable}
}
\end{table*}

\clearpage{}

\section{Conclusion}
We have utilized the GraphLab parallel machine learning framework to efficiently and rapidly  implement multiple collaborative filtering algorithms. having a fast way of testing multiple model settings allowed us for efficient blending of multiple algorithms together. We have further introduced a novel algorithm called MFITR for accounting for item taxonomy. Using fast multicore implementation of multiple algorithms as well combining solution of our  MFITR algorithm allowed us to achieve the $5^{th}$ place at track 1 of the ACM\ KDD CUP\ 2011 contest.



\label{sec:exp}
\section*{Acknowledgement}
D. Bickson would like to thank Liang Xiong from Carnegie Mellon University for his extensive help in implementing the BPTF algorithm \cite{SDM10} in parallel.
D. Bickson would like to thank Joel Welling from Pittsburgh SuperComputing Center for his great support of utilizing BlackLight Supercomputer, generously made available by grant "A Very Large Shared Memory System for Science and Engineering"
OCI-1041726 (PI Levine). Carnegie Mellon work was supported by grants  ARO MURI W911NF0710287, ARO MURI W911NF0810242, NSF Mundo IIS-0803333, NSF Nets-NBD CNS-0721591 and ONR MURI N000140710747.
\bibliographystyle{abbrv}
\bibliography{kdd-cup-workshop}
\end{document}